\documentclass[10pt,journal,compsoc]{IEEEtran}

\usepackage{ifpdf}
\ifpdf
  \pdfoutput=1\relax                   
\else
\fi%

\ifCLASSOPTIONcompsoc
  \usepackage[nocompress]{cite}
\else
  \usepackage{cite}
\fi

\ifCLASSINFOpdf
  \usepackage[pdftex]{graphicx}
  \DeclareGraphicsExtensions{.pdf,.jpeg,.png}
\else
  \usepackage[dvips]{graphicx}
  \DeclareGraphicsExtensions{.eps}
\fi

\graphicspath{{figures/}{pictures/}{images/}{./}} 
\usepackage{comment}
\usepackage{amsmath,amssymb} 
\usepackage{color}
\usepackage{capt-of}
\usepackage{cuted}
\usepackage{xcolor}
\usepackage{hyperref}
\usepackage{subcaption}
\usepackage{wrapfig}

\newcommand{\netvis}{\textit{Net2Vis}}

\begin{document}
\title{
  Net2Vis -- A Visual Grammar\\
  for Automatically Generating Publication-Tailored\\
  CNN Architecture Visualizations
}

%

\author{Alex~Bäuerle,
        Christian~van~Onzenoodt,
        and Timo~Ropinski
\IEEEcompsocitemizethanks{\IEEEcompsocthanksitem All authors were with the Visual Computing Group at Ulm University\protect\\
E-mail: see \url{https://a13x.io}}
}

\markboth{}%
{Bäuerle \MakeLowercase{\textit{et al.}}: Net2Vis}
%



\IEEEteaser{
    \centering
    \includegraphics[width=\linewidth]{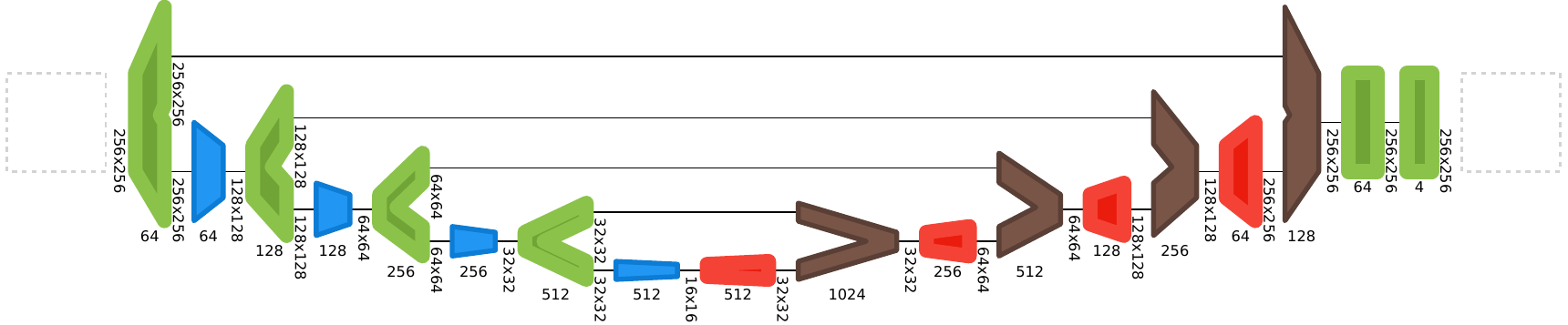}
    \includegraphics[width=0.7\textwidth]{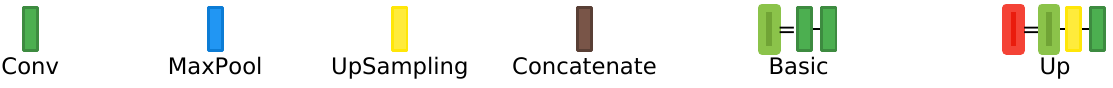}
    \setcounter{figure}{0}
    \captionof{figure}{
      Visualization of a U-Net variant. It was automatically generated using our approach based on the Keras code describing the architecture. Data flows from left to right. Glyphs represent layers or aggregates, while lines represent connections. Glyph widths communicate feature size, while heights communicate the spatial resolution. Both values are also given through labels, while dashed boxes on the left and right serve as placeholders to provide input and output samples. The legend communicates layer types and the composition of aggregates.~\label{fig:teaser}}
}

\IEEEtitleabstractindextext{%
\begin{abstract}
  To convey neural network architectures in publications, appropriate visualizations are of great importance.
  While most current deep learning papers contain such visualizations, these are usually handcrafted just before publication, which results in a lack of a common visual grammar, significant time investment, errors, and ambiguities.
  Current automatic network visualization tools focus on debugging the network itself and are not ideal for generating publication visualizations.
  Therefore, we present an approach to automate this process by translating network architectures specified in Keras into visualizations that can directly be embedded into any publication.
  To do so, we propose a visual grammar for convolutional neural networks (CNNs), which has been derived from an analysis of such figures extracted from all ICCV and CVPR papers published between 2013 and 2019.
  The proposed grammar incorporates visual encoding, network layout, layer aggregation, and legend generation.
  We have further realized our approach in an online system available to the community, which we have evaluated through expert feedback, and a quantitative study.
  It not only reduces the time needed to generate network visualizations for publications, but also enables a unified and unambiguous visualization design.
\end{abstract}

\begin{IEEEkeywords}
Neural networks, architecture visualization, graph layouting
\end{IEEEkeywords}}

\maketitle

\IEEEdisplaynontitleabstractindextext
\IEEEpeerreviewmaketitle

\IEEEraisesectionheading{\section{Introduction}\label{sec:introduction}}

\begin{figure}[t]
  \centering
  \includegraphics[width=1.0\linewidth]{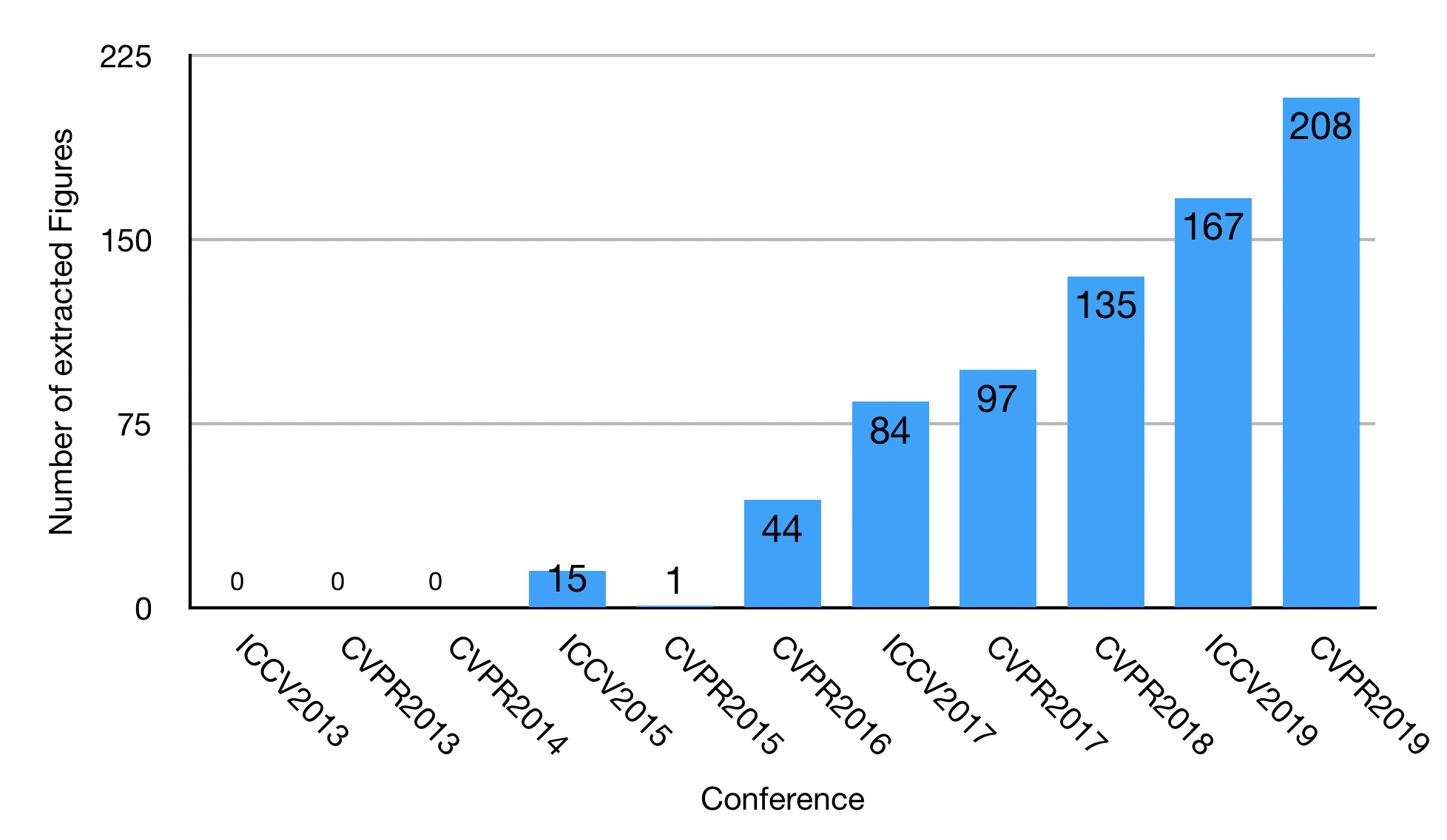}
  \caption{\label{fig:trend}
    Number of CNN architecture figures we extracted from all ICCV and CVPR papers between 2013 and 2019.
    We searched for pages of papers containing figures and the words \emph{figure} and \emph{architecture} in the same line to extract these.
    Then we manually filtered them to obtain only neural network architecture visualizations.
  }
\end{figure}

\IEEEPARstart{P}{apers} utilizing CNNs are published on a daily basis.
An essential aspect of all these publications is to communicate the used or developed network architecture.
Accordingly, a rising number of architecture visualizations can be observed from year to year (see \autoref{fig:trend}).
Authors, who often may lack visualization expertise, mostly use handcrafted, non-standardized visualizations.
As a consequence, generating visualizations takes significant time, and authors often employ suboptimal visual encodings that are sometimes even inaccurate or erroneous.

We argue, as backed by our expert questionnaire (see~\autoref{sec:evaluation}), that the time invested in suboptimal visualizations would be better used to improve training results.
Nevertheless, such abstract visualizations are generally considered to be of great importance.
Therefore, automated approaches that obey to a common visual grammar are required.
We argue that, ideally, such a visual grammar should be informed by three factors: current practice, expert demands, and visualization expertise.
Accordingly, we have analyzed properties of existing architecture visualizations, which we scraped from all ICCV and CVPR papers published between 2013 and 2019 -- which led to a pool of 751 such visualizations.
ICCV and CVPR are prime conferences on machine learning for vision-related tasks and, thus, reflect the great need for such automated visualization approaches.
Additionally, we contacted authors of highly cited papers encompassing architecture visualizations, in order to assess their demands.
Last but not least, we brought in established rules from the data visualization literature to inform our visual grammar.
Based on this, we propose the first method to automatically generate abstract, publication-tailored visualizations of complex, modern CNN architectures, obeying a unified visual grammar, which we refer to as~\netvis.
To this end, we make the following three main contributions:

\begin{enumerate}
  \item We propose a set of requirements for effectively communicating neural network architectures, based on expert feedback and the analysis of existing visualizations.
  \item Based on these requirements, we propose a new visual grammar for CNN architecture visualizations, which we make available via an online platform that transforms Keras code into visualizations tailored to the use in publications.
  \item We release a data set of 751 neural network architecture visualizations, which we have extracted from all papers published at ICCV and CVPR between 2013 and 2019.
\end{enumerate}

\autoref{fig:teaser} shows an example visualization of a U-Net variant generated using our approach.
To evaluate our approach, we conducted both a quantitative user study and a qualitative usability evaluation.
The obtained results indicate that our techniques are beneficial for creating and reading CNN architecture visualizations, which is important for broad acceptance and unambiguous CNN architecture communication.
Our techniques can be used in the form of an online platform at: \url{https://viscom.net2vis.uni-ulm.de}.

\section{Related Work}

Handcrafted visualizations are part of many research papers that use neural networks in their publications~\cite{noh2015learning,teney2016learning,jiang2017super,badrinarayanan2015segnet}.
However, they differ greatly in their visual appearance, which complicates transferring knowledge between them, e.g.,~\cite{strezoski2017plug,long2015fully,nalbach2017deep}.
In addition, they sometimes contain errors, as can be observed in work done by Henzler et al.~\cite{henzler2017single}, where visual glyph encoding and glyph labeling diverge.
Thus, automatically visualizing network architectures to convey their underlying ideas is an extensive field of research.
In the following, we divide related research into approaches for debugging and investigating network architectures, and approaches targeted towards communicating these.
\\
\noindent\textbf{Debugging approaches.}
Demonstrating the importance of visualization for the field, most deep learning frameworks, such as Tensorflow~\cite{abadi2016tensorflow} with TensorBoard~\cite{wongsuphasawat2018visualizing}, Caffe~\cite{jia2014caffe} with Netscope~\cite{netscope}, and also Keras~\cite{chollet2015keras}, directly provide visualization toolkits.
%
All of these are clearly designed for online use.
They are based on vertical layouts for detailed visualizations including all layers and parameters and provide some information only on interaction.
Their consumption of visualization space and required user interaction are perfect for debugging the network architecture, but it renders them inapplicable for use in publications.
\\
Network visualization tools similar but unrelated to these frameworks such as ANNvisualizer~\cite{tudor18ann} and Netron~\cite{netron} suffer comparable shortcomings.
Their glyphs do not convey any information apart from layer type, whereby additional information is displayed by overlaying textual annotations on top of the used glyphs.
Additionally, their vertical layout, along with spacing between layers makes even small networks appear relatively large.
\\
Another interesting visualization approach along this line was presented by Wang et al.~\cite{wang2019visual}.
Here, the focus is on comparing different neural network architectures.
The approach can be used to identify differences in neural architecture design, compare the number of parameters, and draw conclusions for one's own architecture choice.
However, while this approach allows for in-depth comparisons through interactive visualizations, it is not designed to convey network architecture details in a compressed, static way.
\\
\noindent\textbf{Communication approaches.}
Some visualizations convey neural network architectures to explain their functionality to novices~\cite{crowe18neurovis,smilkov2017direct,harley2015interactive,kahng2019gan,liu2018analyzing}, or are targeted towards analyzing what a network has learned~\cite{liu2017towards,zeng2017cnncomparator,bruckner2014ml,kahng2018cti}.
These visualizations clearly fulfill their purpose to support education or interpretability, but are not designed for use in publications.
They all display basic network architectures limited to a specific use case and are not generalizable to more complex architectures.
\\
One visualization technique that is specifically targeted towards use in scientific papers is Drawconvnet~\cite{drawconvnet}.
Convnet-drawer~\cite{convnet_drawer}, which builds on the aforementioned, provides such visualizations, and even allows visualization generation from source code.
Similarly, NN-SVG also claims to create publication-ready network visualizations~\cite{nn-svg}.
While these techniques can be used for small and simple networks, they all face major problems.
First, they do not scale to modern, large network architectures since no aggregation technique is implemented.
Second, they visualize layer connections simply by placing the layers from left to right, which means that parallel network parts cannot be represented.
Additionally, in Drawconvnet and NN-SVG, users have to invest the time to rebuild their network architecture to obtain visualizations.
\\
The surveys by Hohman et al.~\cite{hohman2018visual} and Yuan et. al.~\cite{yuan2020survey} discuss many of these graph visualization techniques.
One important downside of all currently available approaches is that they struggle to visualize large networks in a compact way.
Thus, it remains an open challenge to generate publication-tailored visualizations, despite the existence of the visualization systems described above.
Current state-of-the-art visualizations~\cite{netron,netscope,wongsuphasawat2018visualizing} allow to inspect operations in great detail.
However, these visualizations lack abstractions to make the general network structure comprehensible at a glance.
For a demonstration of this problem, see our supplementary material, which contains a comparison between Netron, Netscope, TensorBoard, and our approach.
Other visualization techniques that aim at providing publication-ready visualizations cannot handle modern network architectures~\cite{nn-svg,convnet_drawer,drawconvnet} and lack important features requested by experts.
Thus, in research papers, these complex networks are usually simplified and drawn manually~\cite{henzler2017single,badrinarayanan2015segnet,nalbach2017deep}.

Besides the extra time effort related to this manual drawing process, the field lacks guidelines to create such visualizations, as is observable in our review of papers from CVPR and ICCV.
Some properties in existing visualizations are ambiguously interpretable, e.g., where downsampling happens, and, for the lack of a common visual grammar, knowledge can hardly be transferred between different publications.
Therefore, we propose a novel visualization technique for abstract architecture visualizations that are optimized for use in scientific publications, where display-space is limited and interaction is impossible.

\section{Design Requirements Abstraction}\label{sec:requirements} 
We argue that for abstract CNN visualizations, both model properties and layer properties need to be visualized.
Model properties are important for the layout and arrangement of the network architecture, while layer properties visualize parameters of individual layers, or groups of layers.
In the following sections, we describe how we used the analysis of current practice and insights from visualization research to inform our design proposition for CNN architecture visualizations.

\noindent\textbf{Data collection.} To support this analysis, we interviewed multiple ML practitioners and reviewed 751 figures of neural network architectures extracted from papers of all CVPR and ICCV conferences between 2013 and 2019.
This data was gathered by crawling~\url{http://openaccess.thecvf.com/} for all 7988 main conference papers using scrapy~\cite{scrapy}.
We then filtered for neural network architecture visualizations and extracted all 1168 pages that contain a figure and had a line of text containing both the words \emph{figure} and \emph{architecture} using PyPDF2~\cite{pypdf2}.
We used pdffigures2~\cite{clark2016pdffigures} to then extract the figures from these pages.
This yielded 1027 images which we manually coded with respect to their visualization design choices.
Since some of the used tools did not work perfectly, we had to delete images that did not contain, or contained corrupted versions of architecture visualizations, leaving a total of 751 figures.
To our knowledge, this is the first data set of such visualizations.
The visualization properties we analyzed are italicized in the following sections and listed in the data set we release alongside this paper.
The following observations are based on both, expert feedback and this figure analysis.
When based on figure analysis, we indicate how many of the analyzed papers used a certain encoding as a percentage value.

\subsection{Model Properties}
In the following, we discuss which model properties of a neural network need to be communicated to quickly assess relevant architectural decisions within a neural network, as derived from the collected data.
\\
\noindent\textbf{Layout.}
Maybe the most important factor when visualizing neural network architectures is the interconnection of layers as it communicates the order in which the computation graph is executed.
At the same time, there is also limited space for publication figures, which is to be taken into account when designing such visualizations.
Thus, one has to find a layout that clearly communicates the order of computations, while also not wasting too much space for a single publication figure.
\\
\noindent\textbf{Connections.}
A consistent layout helps to resolve the order in which the network graph is traversed, but there is still important information missing, namely, connectivity.
Without connectivity information, it is not clear which layer in a network routes data to which following layers.
Especially if network graphs contain parallel execution steps, simply laying out the network layers in a consistent manner will not always resolve which layers actually interact with each other.
Thus, communicating which layers are connected in the computation graph is essential.
\\
\noindent\textbf{Aggregations.}
When thinking about visualizing modern network architectures in publications, space and complexity is a major point of concern.
It is a well-known fact in the visualization literature that hierarchical~\emph{aggregation} can help to simplify visualization designs~\cite{elmqvist2010hierarchical}.
Following this, as networks get more complex, authors manually aggregate layers to make their architectures fit on one page (63.4\%).
Here,~\emph{legends} sometimes indicate which layers are aggregated (15.2\%).
Both of these numbers were rising over the years, as shown in our supplementary material.
We, thus, argue that communicating modern network architectures requires a way of aggregating layer glyphs to create overview visualizations.
\\
\noindent\textbf{Omission.}
Another way of simplifying displayed network architectures is to just omit layers that are not important to convey the general idea of the network architecture.
This is supported by our expert feedback, which indicates that simplification is a major feature for visualizations of neural network architectures.
\\
\noindent\textbf{Input and Output Samples.}
Several of the network visualizations incorporate~\emph{input or output examples} (73.1\%).
However, samples are mainly useful for image or shape related tasks and do not provide additional information concerning the network architecture.
We thus argue that such samples can help for some application areas, while they should not be presented for others.

\subsection{Layer Properties}
Layers are the building blocks of the computation graph that defines any network architecture.
Thus, visualizing properties that parametrize these layers is important to convey the structure and architectural decisions of said network.
In the following, we discuss important layer properties and explain which of those should be visualized for obtaining a general overview of a network architecture.
\\
\noindent\textbf{Layer Type.}
We consider the~\emph{layer type} to be the most important layer-specific variable.
Together with the model properties, it already helps greatly in determining the functionality of a neural network.
It is thus important to encode layer types as a visually prominent variable.
\\
\noindent\textbf{Spatial Resolution.}
Next, to determine the functionality of a neural network architecture, it is important to be able to follow the transformation of data into or out of the latent space, which is often referred to as the change of spatial resolution.
Thus, the dimensionality of data is another variable to be considered when visualizing neural network architectures.
While only slightly more than half of the surveyed visualizations encode the \emph{spatial resolution} (53.7\%), expert feedback, and the fact that this variable can be encoded within the layer glyphs, suggests that visualizing it brings great benefit at no cost.
We thus advocate for always providing information about spatial resolution.
\\
\noindent\textbf{Feature Channels.}
All previously mentioned attributes apply to almost all types of neural networks.
Feature channels, on the other hand, are mainly important for CNNs.
Thus, most visualizations do not at all encode the number of feature channels (56.7\%).
Since feature channels match the variable type of the spatial resolution, and since they are tightly coupled across the network, they are often viewed in combination.
Thus, they support the assessment of data transformation, and our evaluations surfaced that their display is important for architecture visualizations of CNNs.

\subsection{Properties not to be Visualized}

We propose a set of properties to be visualized for providing an overview of a network architecture.
However, there are other properties we explicitly advise to omit in non-interactive visualizations of neural network architectures.
\\
\noindent\textbf{Kernel size.}
~\emph{Kernel sizes} can be found in many visualizations as textual descriptions of layers (24.4\%) or as encodings in the layer glyphs (5.0\%), but are not encoded in most of the visualizations (73.5\%).
When analyzing why kernel sizes are not displayed in most visualizations, two factors were apparent.
First, kernel sizes often are consistent across multiple layers leading to repeated information.
This is in contrast with the request for reduced complexity by domain experts (see~\autoref{sec:evaluation}).
However, the biggest problem with them is that they are in stark contrast with layer aggregations, which are important to reduce the complexity of network visualizations.
For aggregations, there is no such thing as one kernel size as it may differ for layers contained in them.
Additionally, as aggregations are reused, kernel sizes may be different again.
We aim at reducing repetition and embracing aggregation.
Thus, we propose not to display kernel sizes in overview visualizations for neural networks.
This is in line with the expert feedback of domain experts, e.g.~\emph{I'm very interested in great visualization tools that emphasize function and architecture over the kind of "obtuse prettiness" [...]}.
\\
\noindent\textbf{Additional layer properties.}
Neural network layers contain many more features such as weights, strides, padding, and others.
Yet, most of them only exist for certain layer types and are thus rarely communicated.
Some printed visualizations include features such as activation maps, e.g.,~\cite{strezoski2017plug}, or receptive fields, e.g.,~\cite{shi2017ssgan}, but they are only provided for specialized use cases.
In contrast, we argue that for obtaining a general network architecture overview, such features are not necessary.
This is supported by our expert feedback and in line with our goal of providing an abstract overview of the network architecture, where detailed information can be obtained by reading the publication it is contained in.
\\
\noindent\textbf{Dimensionality.}
Three-\emph{dimensional} visualizations are helpful if the reader's task includes shape understanding, but less so for any relational task~\cite{munzner2014visualization,st2001use}.
Thus, the relation of layers and their spatial position would suffer from being visualized in 3D, while the benefit of using three-dimensional layer glyphs would only be to resemble the shape of data in case it is three-dimensional.
Another important argument against three-dimensional visualizations in publications is that the viewpoint cannot be changed.
Contrary, exploration is essential for utilizing the benefits of three-dimensional visualizations~\cite{munzner2014visualization,cockburn2000evaluation}.
About half of the analyzed network visualizations display layers in 3D (50.2\%), however, this added dimension does not convey additional information for most of them.
Since the reduction of spatial resolution is almost always applied to all spatial dimensions equally, the third dimension is not needed for the visualization of such changes.
For those reasons, we advise not to visualize layer glyphs in 3D.

\subsection{Summary}

Based on these requirements, we propose to use the following properties in any visualization that aims at providing a general overview of a CNN architecture in non-interactive visualization environments:

\noindent\textbf{Model Properties}
\begin{enumerate}
  \item Layout
  \item Connections
  \item Aggregations
  \item Omission
  \item Input and Output Samples (for some tasks)
\end{enumerate}

\noindent\textbf{Layer Properties}
\begin{enumerate}
  \item Layer Type
  \item Spatial Resolution
  \item Feature Channels (for convolutional layers)
\end{enumerate}

We, thus, elaborate on \emph{what} to visualize in this section, defining the dimensions of our proposed design space.
This assessment is based on our analysis of figures extracted from the top conferences in the field, as well as expert feedback both before and during the development of our approach.
It is strictly tailored towards conveying the overall idea of a network architecture, and does not cover cases in which specific features of a network are to be communicated.
Therefore, we also provide guidelines for which features not to visually encode for abstract architecture visualizations, thus preserving simplicity and preventing the need for interaction.
By always visualizing the aforementioned parameters in the same way, users can transfer knowledge between different visualizations.
We thus advise not to render the mapping of variables customizable.
Instead, we propose a unified visualization design for these parameters.
In the following sections, we map these visualization properties (i.e., \emph{what} to visualize) to specific visual encodings (i.e., \emph{how} to visualize them), describing the proposed design space inspired by similar design space descriptions~\cite{brehmer2016timelines,felix2017taking}.

\section{Visualization of Model Properties}\label{sec:graphlayout}

Based on our assessment of important visualization aspects for communicating CNN architectures as described in the previous section, we now explain \emph{how} we propose to visually encode the model properties.
For these global properties, the design space consists of the aforementioned dimensions, i.e., Layout, Connections, Aggregations, Omission, and Data Samples, which we describe in the following subsections.

\subsection{Layout}
The representation of this design space dimension mainly concerns the spatial arrangement of layers which can be vertical or horizontal, and in any direction.
Most investigated neural network visualizations layout their layers from left to right (81.6\%).
Our figure analysis as well as the collected expert feedback indicate that for publications, this layout is preferable over vertical layouts, which are often used in online tools.
It does not only preserve the reading direction of western cultures, but visualizations also nicely fit across the width of a page.
Furthermore, perceptual rankings indicate that position best encodes ordered data~\cite{mackinlay1986automating,causse2009physiological,heer2010crowdsourcing}, such as the order of network layers.
Follow these insights, and the fact that space taken up by publication figures is important, we propose to employ a narrow horizontal layout, in which parallel execution steps of the network are stacked vertically on the same horizontal position.
\\
To layout CNN graphs, we propose to use the network simplex algorithm~\cite{gansner1993technique} which is explicitly targeted towards drawing directed rank-based graphs.
The rank-based nature of this algorithm perfectly fits our use case in which parallel layers are to be placed at the same x-coordinate, and sequential parts of the network tend to be drawn on the same vertical level.
Using an algorithm that only layouts series-parallel graphs is not an option, since Keras operations are not restricted to those (e.g., $[a \rightarrow b, a \rightarrow c, b \rightarrow d, c \rightarrow d, d \rightarrow e, b \rightarrow e]$). 

\subsection{Connections}
\begin{figure}[t]
    \centering
    \begin{subfigure}[t]{0.45\linewidth}
      \centering
      \includegraphics[width=0.5\linewidth]{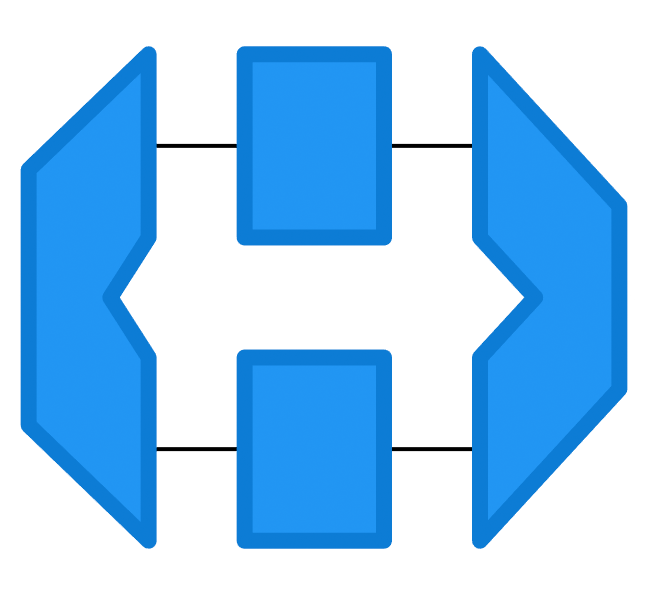}
      \caption{\label{fig:multicon}
        Data-paths
      }
    \end{subfigure}
    \begin{subfigure}[t]{0.45\linewidth}
      \centering
      \includegraphics[width=0.5\linewidth]{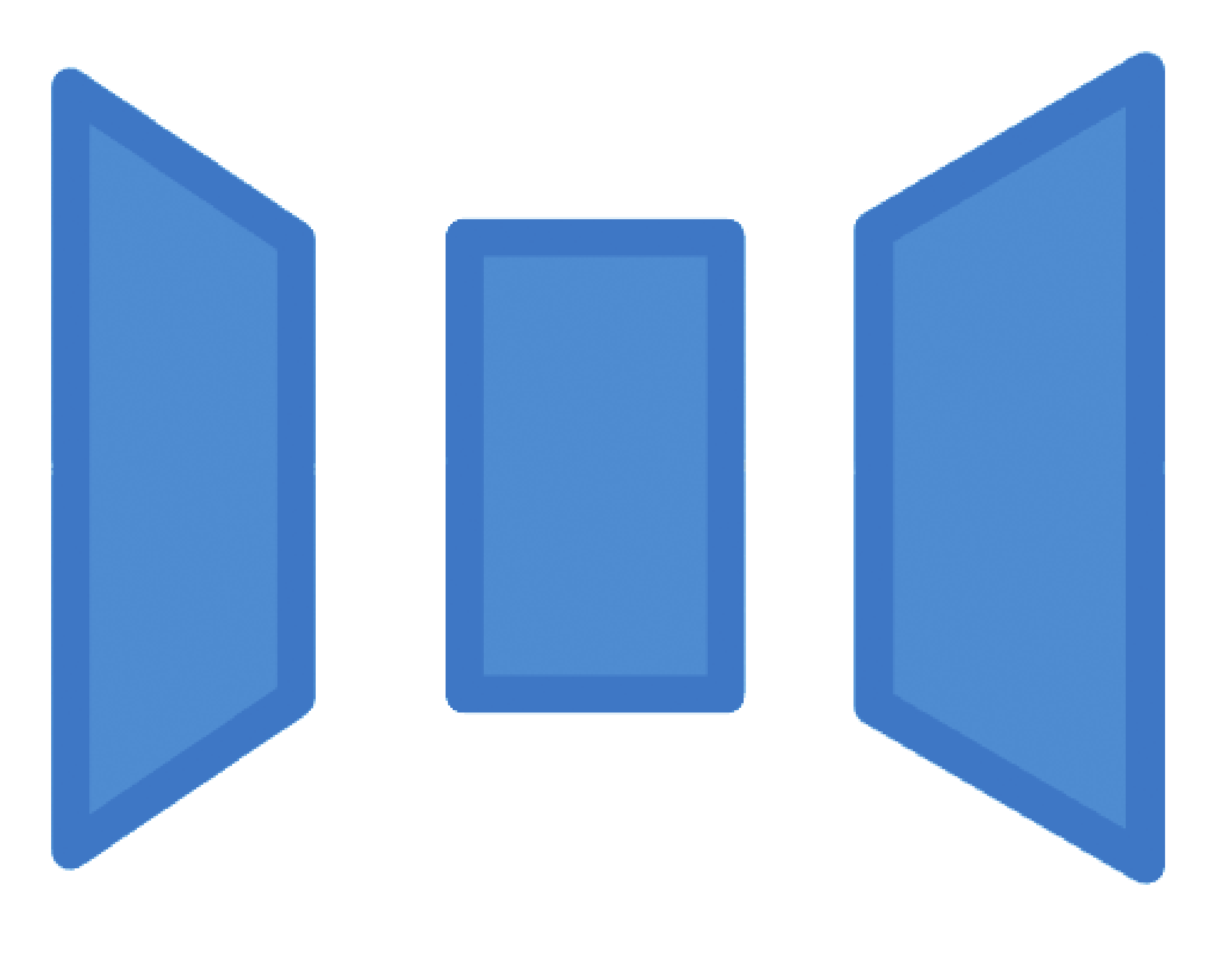}
      \caption{\label{fig:glyphs}
        Dimensionality
      }
    \end{subfigure}
    \begin{subfigure}[t]{0.45\linewidth}
      \centering
      \includegraphics[width=0.4\linewidth]{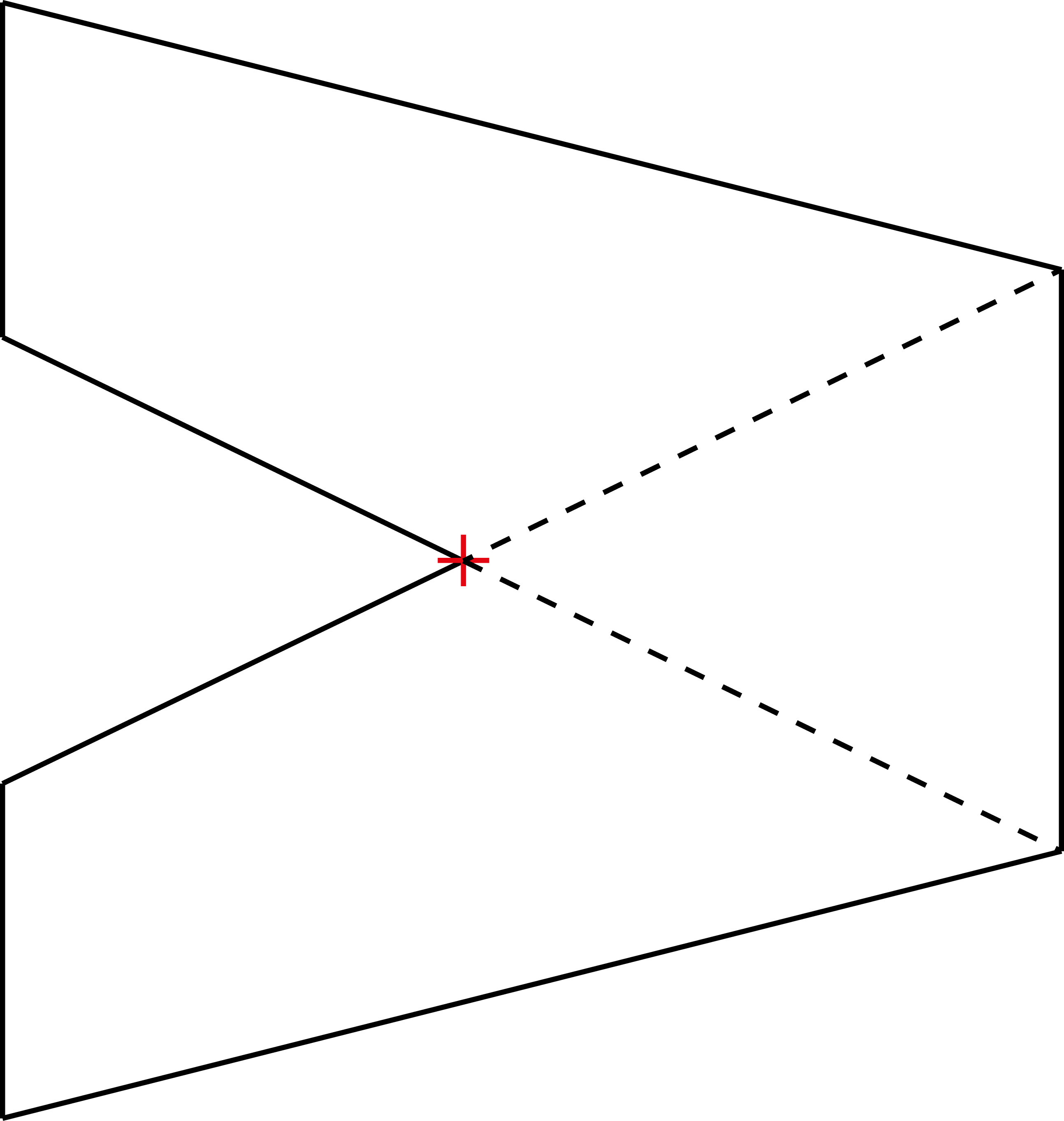}
      \caption{\label{fig:multi-in-calc}
        Handles
      }
    \end{subfigure}
    \caption{
      (a): Two simultaneous data-paths. The layer displayed on the left has two outgoing connections, while the layer displayed on the right fuses these paths back together.
      (b): Left: layer that reduces the spatial resolution; Center: same spatial resolution for input and output; Right: a layer that increases the spatial resolution
      (c): Multi-handled glyphs are always connecting corners on the input side with corners on the output side, here depicted by a dotted line.
    }
\end{figure}

For this dimension of the design space, possible representations are an implicit connectivity without explicit visualizations and different forms of connecting lines.
In existing figures, connections between layers are visualized either using lines (73.4\%) or by simply placing layers next to each other.
Some visualizations additionally add arrowheads to clarify the~\emph{direction} of data flow (65.9\%).
Following most visualizations, we also propose to use lines as connections between layers to emphasize the graph structure of neural networks.
However, we propose not to add arrowheads to these connections, as the forward data flow direction is uniformly left-to-right in our visualizations.
\\
Many architectures contain~\emph{skip connections} visualized by lines between distant layers (55.9\%, rising over the years, see supplementary material).
Displaying splits in the execution graph only through lines has the negative implication that size-related attention bias is induced~\cite{proulx2008biased,proulx2010size}.
Thus, we propose a glyph design that prevents such issues.
Whenever a layer has multiple outgoing or incoming connections, we modify the glyph that represents it as shown in~\autoref{fig:multicon}.
This way, there might be multiple ends on the left or the right side of the glyph each having the same visual prominence.
At the same time, splits and joins of the data flow, which are important features of the architecture, are highlighted.
A visual representation of such multi-handled glyphs is illustrated in~\autoref{fig:multi-in-calc}.
\\
One might think that this layer shape induces problems with edge crossings.
However, edge-crossings are uncommon to neural network architectures.
We have only found planar network graphs which consequently can be laid out to avoid crossing edges.

\subsection{Aggregation}
\begin{figure*}[t]
  \centering
  \includegraphics[width=\linewidth]{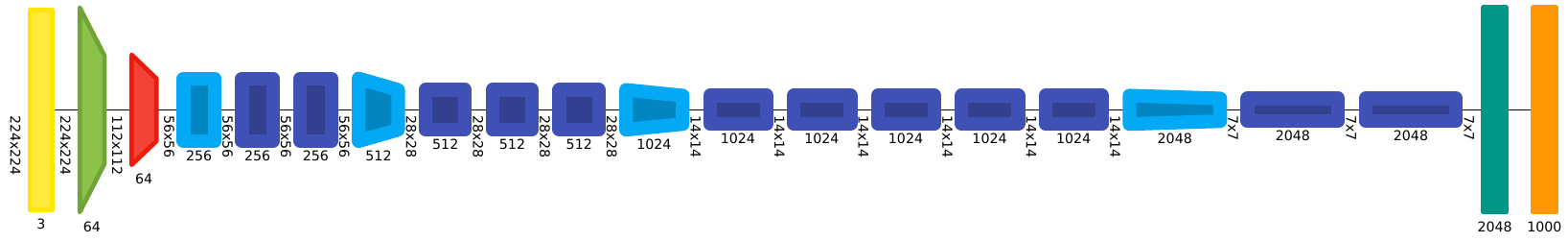}
  \includegraphics[width=.7\linewidth]{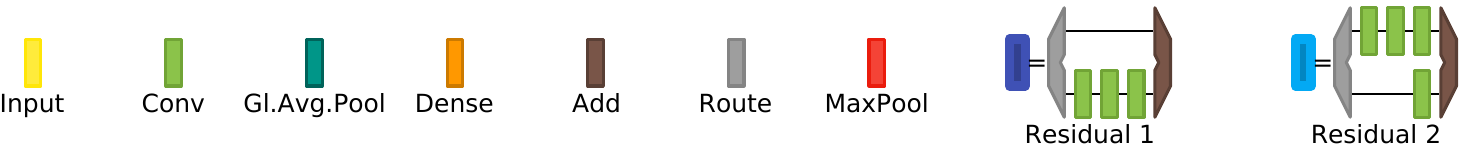}
  \caption{\label{fig:resnet}
    With ResNet50, the removal of activation and batch-normalization layers from the visualization, which do not add information about the network structure, along with the repetition of residual blocks in ResNet allows us to reduce the number of glyphs that need to be drawn from 177 to just 21, even though we added routing layers to clearly define the beginning of a residual block.
  }
\end{figure*}

The aggregation of multiple layers is another dimension in our design space.
Here, representations can be no aggregation of layers, vertical stacking, or replacement with a group-representing glyph.
As modern networks become increasingly complex, the aggregation of layers is inevitable.
Such aggregations can contain many layers at once, and even parallel paths, which is why we do not employ a stack-based visualization.
Instead, we adopt the paradigm of providing ways to aggregate multiple layers and replacing them with a new, single glyph.
As this removes direct insight into the content of aggregations, we resolve them in a legend below the network graph.
Furthermore, domain experts and users frequently requested a visual separation of aggregations.
Thus, in our approach, their border is drawn thicker, and their color scheme is inverted (lighter border than center).
\\
\noindent\textbf{Aggregation constraints.}
As aggregations substitute all occurrences of selected layer sequences throughout the graph, only parts of the network graph that can be represented by one sequential layer can be aggregated.
We, thus, restrict aggregation to sequentializable segments.
This way, no deformed aggregation layers can occur, where two outputs or inputs might differ in their spatial resolution.
This ensures visual consistency such that layers always manipulate the data the same way for all of their connections and that all connections of each layer end on the same horizontal level in the graph.
\\
\noindent\textbf{Automatic aggregation.}
To generate aggregations, the layers to be aggregated can either be selected by the user or more conveniently be selected automatically.
To obtain automatic aggregations, we propose to analyze all sequential parts of the network.
We then search for recurring sequences of layers.
The most frequent of these sequences is then assumed to be the preferred aggregation.
\\
\noindent\textbf{Interacting with aggregations.}
We argue that visualization designers should be able to remove or temporally deactivate aggregations of the network, which expands abstracted layers back to their initial layout.
Deactivated aggregations are preserved in the legend for later reuse to be able to explore the visualization without losing information.
To visually convey the state of aggregation, we propose to draw active ones with a dark outline and black description text, while outlines and text are drawn in light gray for inactive aggregations and layer types that are hidden by the user.
This way, the effect of different aggregation levels can be explored while preserving all aggregation information.
\\
\noindent\textbf{Split layers.}
While there are dedicated layers to fuse computation paths, routing the data to multiple outputs is done implicitly.
Thus, it is possible, that, e.g., an activation layer feeds data to two different paths.
Visually, this is a problem as splits and merges of the computation graph are often seen as blocks and frequently get aggregated by the user.
Activations are orthogonally seen as the end of a computation group rather than a start of one.
For such dedicated groups, we argue that the user should be able to add special routing layers.
This way, they can clearly communicate the special role of such multi-path aggregations while also assigning more importance to data path splits, e.g., as shown in~\autoref{fig:resnet}.

\subsection{Omission}
In the computation graph, any function that has been added by the developer is considered a network layer.
However, these graphs can be defined at different levels of detail (e.g., activation within layer or separately).
Thus, in our approach, this graph can be thinned by the user to better convey the underlying architecture rather than each individual computation step by hiding individual layer types entirely.
Showing a graph overview, and then allowing the user to filter it is in line with Shneiderman's mantra~\cite{shneiderman2003eyes}.

\subsection{Input and Output Samples}
As described in the property analysis of~\autoref{sec:requirements}, input and output samples may, for some networks, be helpful.
Directly integrating such samples would, however, require the user to provide training or testing data, and thus interfere with the automatic nature of our visualization design.
We thus propose not to include them directly in any programmatic CNN architecture visualizer.
Instead, we suggest to optionally provide placeholders for input and output samples which users can replace during post-processing with actual samples.

\section{Visualization of Layer Properties}\label{sec:properties}

In the following, we describe the design space dimensions of the visual layer encoding we propose based on the design discussion presented in~\autoref{sec:requirements}, in other words, \emph{how} to visualize layer properties.
Our visualization design supports the direct encoding of layer type, spatial resolution, and the number of feature channels.
By reducing the visualization space to these three variables, we are able to encode all of them in simple glyphs that represent the layers of the network architecture to be visualized.
For the spatial resolution and the number of feature channels, percentages only consider the 464 visualizations that contain convolutional layers, as we explicitly tailored our glyph design to work well for CNNs.

\subsection{Layer Type}
The design space dimension for the layer type can take the representations of textual display, color-coding, and shape-encoding. 
Most visualizations only use textual descriptions (52.7\%), or text along with glyph color (14.6\%) to convey layer types.
However, it can be repetitive to encode the layer type as text below each layer.
We, thus, propose to provide this textual encoding optionally (see~\autoref{fig:3d}) while not being displayed in the default setting.
The layer type is a categorical attribute and consequently best visualized using a channel that is optimized for such data~\cite{munzner2014visualization,ware2012information}.
In Mackinlay's ranking~\cite{mackinlay1986automating}, color ranks just behind position, which we already employ to communicate the data flow of the network.
Perception research also shows that color is a visually dominant channel~\cite{christ1975review}.
Thus, we propose to use a color encoding to convey the layer types in the visualized network.
\begin{figure}[b]
  \centering
  \includegraphics[width=1\linewidth]{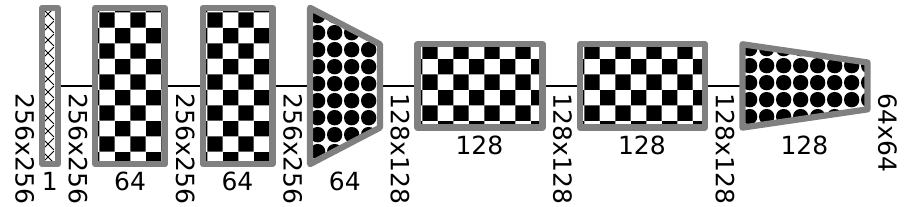}
  \\
  \includegraphics[width=0.7\linewidth]{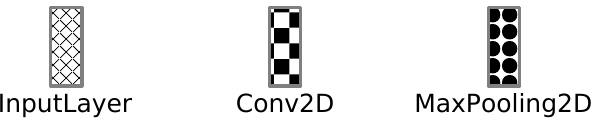}
  \caption{\label{fig:colorblind}
    Accessible encoding of the layer type to also support readers with monochromatic vision and publications without color.
  }
\end{figure}
\\
\noindent\textbf{Color Assignment.}
In our approach, colors for new layers are automatically preset in accordance to one of two alternative approaches.
The first approach is motivated by farthest point sampling.
It finds unused colors in hsv color space by searching for the biggest gap between any two hue values of already used colors.
This is the most functional approach, as it optimizes for color difference.
Unfortunately, it might result in colors that are indistinguishable by color-blind users and unpleasant color choices.
Therefore, the second option for color proposition is palette-based and serves as the default.
We suggest to use two color palettes, one from materialuicolors~\cite{network18materialuicolors} for visually pleasing color mappings, and one adapted to users with trichromatic or dichromatic vision~\cite{wong2011points}.
Additionally, visualization designers should always be able to customize the color that is used to encode a layer type. 
\\
To also make our visualizations accessible to readers with monochromatic vision, and support publications without colored images, we also propose a texture-based encoding of layer types, see~\autoref{fig:colorblind}.
We provide twelve distinguishable patterns that can be extended upon when needed.
\\
\noindent\textbf{Legend Generation.}
Since we use color-coding to differentiate between layer types, a legend that maps these color codes back to layer names is needed.
This legend contains a glyph for each layer type in the network and displays the name of its layer as shown in~\autoref{fig:resnet}.
Based on expert feedback, legend items are sorted from simple to complex.
This complexity is determined by analyzing the dependency-tree of aggregations, whereas nested aggregations are seen as more complex.

\subsection{Spatial Resolution}
The spatial resolution is a design space dimension that might be represented by glyph height, glyph width, glyph color, and textual annotations.
As the spatial resolution is a quantitative and sequential variable, length, angle, slope, and area are the best remaining options for encoding it~\cite{mackinlay1986automating,causse2009physiological,cleveland1984graphical}.
In some visualizations, the spatial resolution is represented by the shape of the layer glyph in combination with textual information (10.8\%) or just textual representations (10.1\%).
However, mostly only glyph shape (32.8\%) is used.
We propose to use height in combination with text as a direct mapping of the spatial resolution, as this does not interfere with our network layout and can be encoded in the glyph design directly.
\\
However, not all visualizations map glyph height in the same way.
Sometimes the height of the downsampling layer already encodes the changed resolution, while in other visualizations, the next layer is first affected by this change.
This ambiguity makes the interpretation of such visualizations hard, as one needs to determine which representation was chosen for each visualization approach.
Furthermore, the transformation of the resolution is determined by multiple parameters (e.g., stride, kernel size, padding).
Thus, the output resolution is a result of the inner working of a layer rather than a fixed parameter.
We, therefore, propose to visualize the spatial resolution as a change within the layer.
To convey the underlying transformation, we set the height on the left edge of the glyph to match the input resolution, while the height on the right edge of the glyph reflects the output resolution, resulting in trapezoid-shaped glyphs, as shown in~\autoref{fig:glyphs}.
This conveys the change of resolution as a result of the mathematical operations within the layer while at the same time removing its ambiguity.
In addition, with this encoding, the relation of input and output dimension from layer to layer as well as across the whole network is clearly visible by horizontally scanning the visualization~\cite{kim2018assessing}.
\\
To draw these glyph shapes, in our approach visualization designers can define a minimal and maximal height for the glyphs. We then obtain the spatial resolution for the input and output tensors for each layer.
The highest and lowest value of all spatial extents gets mapped to the extremes of the user-defined height values.
Values between these extremes are interpolated linearly to convey the actual spatial resolution for the input and output of each layer in the network.
Using these interpolations, each input and output of every layer gets a height value assigned, which maps this important quantitative variable to the vertical length of the glyph ends, as Mackinlay's ranking suggests~\cite{mackinlay1986automating,cleveland1984graphical}.
\\
We found that it is mostly not important to convey the exact spatial resolution by means of textual descriptions, as only 20.9\% do so.
Since many modern architectures further allow inputs to be arbitrarily shaped, e.g.,~\cite{jiang2017super,nalbach2017deep,zhou2017voxelnet}, the spatial dimension is not necessarily fixed at any given layer for many network architectures.
We, therefore, advocate for the option to toggle labels that display the exact spatial resolution between the layers, following our visualization design, in which the resolution is fixed between layers but changes within them.

\subsection{Feature Channels}
Similar to the spatial resolution, feature channels may be represented by glyph height, glyph width, glyph color, and textual annotations.
Textual descriptions (20.0\%), glyph shapes (13.8\%), or a combination of both (9.1\%) are common for conveying the number of feature channels.
The number of feature channels, just as the spatial resolution, represents the important transformation into or from latent space, tightly coupling these two variables.
Thus, we propose to employ a similar visual encoding to convey them, again, mapping a perceptually preferable length parameter, in this case, glyph width, to this variable~\cite{mackinlay1986automating,heer2010crowdsourcing}.
Feature channels are different from the spatial resolution in that they are fixed properties of a layer and not derived from the previous dimensions.
We, therefore, propose to visualize this variable as a direct property of the layer rather than a change within it.
In our approach, the number of feature channels can additionally be represented as text, displayed below each layer.
\\
While the spatial resolution and its change within a layer is represented by the heights at both ends of the layer glyph, combining this representation with the number of feature channels as encoded in the glyph width can reveal important overall information on the processed data.
Together, they present the total amount of data that is processed by the layer.
In our visualization design, this amount of data is implicitly encoded in the area covered by the glyph.
\\
Accordingly, with this glyph design, variable importance is perfectly aligned with Mackinley's ranking~\cite{mackinlay1986automating,causse2009physiological,kim2018assessing}.
At the same time, these glyph shapes fit nicely into the horizontal network layout.
\\
Dense layers only have one intrinsically specified dimension of data.
Since this is a fixed dimensionality, it is more similar to feature channels than the spatial resolution.
Additionally, dense layers are one-dimensional and commonly visualized as vertical chains of neurons.
Thus, for dense layers, the number of neurons is also mapped to the glyph height.
Additionally, for better differentiation of these simpler layer types, we also propose to employ a simpler color-coding, using the same color for the border as well as for the body of the glyph.

For all of these visual encodings, we propose default specifications, such as minimal and maximal width and height of layer glyphs, which can be customized by the user.

\section{Evaluation}\label{sec:evaluation}

To evaluate the proposed concepts, we have visualized multiple well-known architectures with our techniques, gathered expert feedback to inform and evaluate our visualization design, and conducted a quantitative user study. 

\subsection{Application Examples}\label{sec:examples}

\begin{figure}[b]
  \centering
  \includegraphics[width=0.7\linewidth]{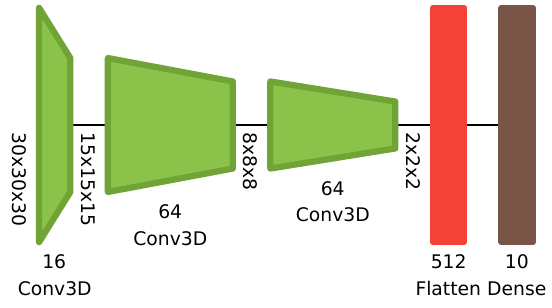}
  \\
  \includegraphics[width=0.7\linewidth]{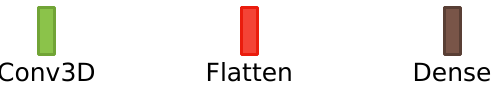}
  \caption{
    \label{fig:3d}
    \netvis~can also be used to generate visualizations of multi-dimensional network architectures.
  }
\end{figure}

To provide the community with means to incorporate our visual encoding, we implemented~\netvis~as an online application which is available at \url{https://viscom.net2vis.uni-ulm.de}.
Here, users can paste Keras code to obtain ready-to-use architecture visualizations and download those as PDF figures for direct use in publications, as well as SVG images, which can be edited in hindsight.
To demonstrate \netvis' capabilities, we applied it to several commonly used network architectures.
~\autoref{fig:teaser} shows a variation of U-Net~\cite{ronneberger2015u} which is frequently used for semantic segmentation.
~\autoref{fig:resnet} shows a visualization of ResNet~\cite{he2016deep} where we show a reduction from 177 to just 21 glyphs through our aggregation techniques.
Finally, in~\autoref{fig:3d}, we demonstrate that we also support multi-dimensional network architectures.
Even more application examples where we show that our techniques can even visualize networks such as InceptionV3~\cite{szegedy2016rethinking} can be found in our supplementary material.
Here we show popular published neural network architectures~\cite{szegedy2016rethinking,long2015fully,howard2017efficient,he2016deep,noh2015learning,ott2018deep,ronneberger2015u,simonyan2014very,nehme2018deep,urban2018deep,he2016identity}, and two network visualizations we used for our own publications and presentations.

\subsection{Comparison to TensorBoard}
TensorBoards graph visualization is based on the computation graph as defined in the program code for the network architecture.
Aggregations in the graph visualizer can, thus, only be made for those nodes that contain sub-nodes.
This can provide more detail as every operation in the graph can be examined.
However, TensorBoard's aggregation approach is far less flexible and not tailored towards publication figures.
For publication visualizations, users typically want arbitrary aggregations of multiple layers without having to specify parent nodes for those in the code first.
Our flexible graph-based aggregation methods support that exact need, in that they allow for tailoring aggregations in a way that best communicates the overall architectural ideas.
This also leads to much smaller and easier to understand figures, which we will elaborate more on in the following evaluations.

\subsection{Expert Interviews}
After the implementation of the first version of~\netvis~was complete, we conducted qualitative interviews with experienced machine learning researchers.
\\
\textbf{Expert selection.}
We used Net2Vis to generate replications of visualizations from published papers.
These visualizations were then emailed together with our questionnaire to the authors of the respective papers (in the following referred to as experts).
In total, we contacted 7 experts in this manner who had novel papers or high citation count and published at different conferences.
Three of those answered our questions, one replied to have other obligations, and three did not get back to us.
Two of the papers from which the respective experts gave feedback are highly cited (i.e., Noh et al.~\cite{noh2015learning}: $>2300$ and Long et al.~\cite{long2015fully}: $>14000$), the third is a more recent publication from 2018~\cite{ott2018deep}.
\\
\textbf{Questionnaire.}
Our questions were designed following Munzner's nested evaluation model~\cite{munzner2009nested}.
Thus, we assessed the need for such automatic visualizations (Q1, Q5), analyzing the threat of targeting a wrong problem.
We also investigated why 3D visualizations are so common (Q3), and asked about our visualization design (Q2, Q4) to evaluate the abstraction and encoding technique, which are the second and third possible pitfalls~\cite{munzner2009nested}.
Our implementation proves interactivity, the fourth possible visualization pitfall~\cite{munzner2009nested}.
We intentionally asked only five questions to keep the time needed to answer our survey relatively low, and thus maximize the chance for responses from these well-known researchers.
However, we encouraged the experts to add any comments to their replies.
\\
\textbf{Feedback.}
All replies were positive and emphasized the importance of such automatic visualizations.
Furthermore, they gave positive feedback on our glyph and graph design.
Their main concerns were scalability to different architectures since they only received visualizations for their papers.
However, as can be seen in~\autoref{sec:examples} as well as the supplementary material,~\netvis~is designed to work with a wide variety and high complexity of convolutional neural networks.
\\
\noindent\textbf{Q1: How much time did you spend creating your figure?}
All experts stated that creating figures of their architecture took too much time.
While initial versions seemed to take about one hour on average, they all noted that they needed multiple iterations.
This supports our claim that this task can be greatly optimized as it takes valuable research and paper-writing time from the experts.
\\
\noindent\textbf{Q2: How do you understand the mapping of number of feature channels and spatial resolution in the visualizations we sent you?}
All three responded that they understand that and how the spatial dimension and feature channels are mapped onto the glyphs correctly.
Thus, the mapping of spatial dimension and feature channels seems comprehensible even though this representation is more abstract than the ones used in their papers.
\\
\noindent\textbf{Q3: Why did you pick a 3D visualization for the layers and which information did you want to convey?}
All experts said this was done since the data was three-dimensional.
None of them conveyed additional information this way.
One expert also admitted that 3D visualizations introduce the problem that layers cannot always be evenly spaced because of occlusion.
All three also noted that this makes the visualization more complex.
Our choice of visualizing the network architecture in 2D was preferred also by these experts.
\\
\noindent\textbf{Q4: What do you think of visualizing the transformation that happens during pooling/unpooling as a transformation of the layer itself (trapezoid glyphs) rather than in-between?}
One expert said \emph{I found many people complained about not drawing in-between relation between pooling/unpooling}, which indicates that this implicit transformation used in the existing visualizations is confusing to the reader.
Another expert mentioned that \emph{the trapezoids seem like a nicely simple way to indicate where and how much downsampling is going on}.
However, he also noted that this is dual to the way AlexNet visualizes network architectures which has been picked up by many researchers.
While it is a valid concern in that readers have to differentiate between these approaches, we think that the mentioned benefits outweigh the downsides which naturally come with adopting a new visualization approach.
\\
\noindent\textbf{Q5: Would you use such a tool for your projects, if available?}
All experts agreed that they would be users of network visualization generators as proposed in this paper.
One expert additionally mentioned that he would still want to have the possibility to modify the visualization to his will which he did not know is possible in~\netvis.
Other remarkable comments that clearly show the need for such automatic visualizations were: \emph{I have been ranting about this for a while and have been waiting for somebody to ask}, and \emph{I've been hoping someone would make better automatic visualization toolkits for a while}.

Based on the comments and free-form texts within the expert feedback, we further refined our visualization design, e.g., by visually separating groups from standard layers through different border styles or increased vertical spacing between parallel execution steps.

\begin{figure}[t]
  \centering
  \includegraphics[width=0.8\linewidth]{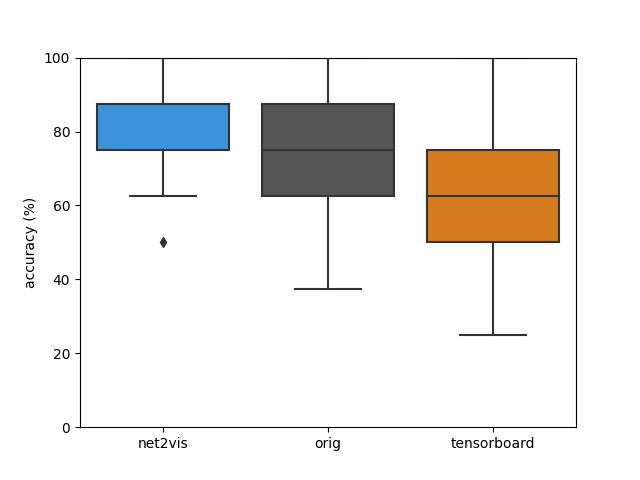}
  \caption{\label{fig:accuracy}
    This plot shows a comparison of accuracies for each condition.
    Accuracies are the mean of all eight questions asked for each network.
    For~\netvis, the median is the same as the upper IQR.
    One can see that our approach led to the best accuracy, even though the other approaches contained questions that participants did not even need to answer.
    We additionally found significant differences between TensorBoard and the other two conditions, indicating that TensorBoard is the worst approach of the three for an abstract display of network architectures.
  }
\end{figure}

\subsection{Quantitative User Study}
To evaluate our final visualization design, we then conducted a quantitative study with 10 participants (3 female, 7 male, $M_{age} =$ 25.9 $SD =$ 2.27).
These participants were recruited in a university setting.
Requirements for participation were that the participants knew what a CNN is, knew what elements CNNs consist of, and knew about feature- and spatial dimensions.
Machine learning expertise of the participants varied between less than half a year and less than five years, which nicely reflects the broad audience which we expect for our visualization design.
Based on internal piloting, we found that our study takes participants around an hour to complete.
We compensated them with a 10 Euro Amazon gift card.
\\
\noindent\textbf{Methods.}
The participants were presented with different well-known machine learning architectures.
Each of these architectures was visualized using three different visualization approaches,~\netvis~, TensorBoard, and a visualization taken from the original publication.
We implemented the network architectures for each paper in Keras, which enabled us to directly export visualizations using~\netvis.
For TensorBoard, we had to manually screenshot and stitch the visualizations as the export functionality in TensorBoard did not work for us.
For each visualization, participants had to answer eight questions by extracting information about the architecture.
These questions included the following tasks: \textit{How many convolutional layers does this architecture contain?}, \textit{What is the maximal feature depth for the convolutional part?},  \textit{What is the minimal spatial resolution of the convolutional part?}, \textit{What are the input dimensions for this network?}, \textit{What are/is the output dimension(s) of this network?}, \textit{How many times does downsampling happen in this network?}, \textit{How many steps are performed to increase the feature dimension?},  \textit{Is this Architecture ``Fully Convolutional``?}.
Participants entered the answers to these questions into a text field and were instructed to answer \textit{-1} if they could not extract the information from the visualization.
Each participant was presented with every network architecture (6) using all three visualization techniques (3), resulting in 18 stimuli presented in a randomized order.
Afterward, participants were presented with each architecture using all three visualization techniques, side by side, and were asked three comparison questions: \textit{Which of the above visualizations contains the most useful information?}, \textit{Which of the above visualizations was easiest to interpret?}, and \textit{Which of the above visualizations did you find most visually appealing?}.
\\
\noindent\textbf{Analysis.}
To analyze the performance of the different visualization approaches, we computed the mean accuracy over all eight questions for each of the presented architectures for each condition (\netvis~vs. TensorBoard vs. handcrafted), as can be seen in~\autoref{fig:accuracy}.
We compared these conditions using Friedman ANOVA and found a significant difference in accuracy for the visualizations drawn using~\netvis~($M$ = 83.75\%, $SD$ = 12.02\%) when compared to the handcrafted ($M =$ 75.83\%, $SD =$ 15.75\%), and TensorBoard ($M =$ 63.13\%, $SD =$ 17.29\%), $\chi^2(2) =$ 38.75, $p <$ .001.
During post-hoc analysis using Nemenyi's test, we found a significant difference between~\netvis~and TensorBoard ($p =$ .001), as well as TensorBoard and the handcrafted version ($p =$ .001).
While the accuracy for~\netvis~was higher compared to the handcrafted version, we could not find a significant effect between these conditions ($p =$ .11).
When analyzing the time our participants took to complete all eight questions for each visualization, we found a significant effect between our conditions.
As with accuracy, TensorBoard showed the worst performance ($M =$ 27.1s, $SD =$ 30s), followed by~\netvis~($M =$ 16.98s, $SD =$ 15sec), and the handcrafted version ($M =$ 13.53s, $SD =$ 8.95s), $\chi^2(2) =$ 17.1, $p <$ .001.
During post-hoc analysis, we found this effect to be significant between TensorBoard vs.~\netvis~($p <$ .05), and TensorBoard vs. handcrafted ($p <$ .001).
However, when comparing~\netvis~and the handcrafted version, we could not find a significant effect between these conditions ($p =$ .23).
\\
For the questionnaire about which visualization technique our participants would prefer with respect to informativeness, interpretability, and design, our technique outperformed the other conditions again.
For informativeness, participants preferred our approach in 86.6\% of cases, while they favored the original version in 13.3\% of cases.
The interpretability of our design was ranked highest as well, as in 75\% of all cases, our approach was favored, whereas the original versions were favored in 25\% of cases.
The design of our approach was favored in 70\% of all cases, against 30\% in favor of the original visualization.
Looking at the individual networks, our approach was rated better or evenly good in all conditions except for the design of U-Net, where 60\% favored U-Net.
Remarkably, across all six conditions and all ten participants, TensorBoard did not get voted as preferable once.
Used visualizations, plots, and raw study data can be found in our supplementary material.

\subsection{Usability Evaluation}
In another evaluation with 16 participants (13 male, 1 female, 2 did not report, $M_{age} =$ 28.06 $SD =$ 4.23), we also evaluated our approach from the view of a visualization designer.
These participants took part in our study right after a one-week full-time deep learning course, where we recruited the participants.
Thus, they were familiar with the underlying concepts.
Participants were given a brief introduction to \netvis~before they had the chance to visualize one of their own architectures.
Then, they filled a questionnaire regarding the system, including a system usability scale questionnaire (\textit{SUS}), and a demographic questionnaire.
The usability analysis through the \textit{SUS} questionnaire resulted in a mean score of 83.44 points ($SD$ = 6.25) indicating \textit{excellent} usability~\cite{bangor2009determining}.

\subsection{Discussion}
The main goal of \netvis~was to introduce a unified design for neural network architectures as a replacement for handcrafted visualizations.
While we could not find a significant effect between \netvis~and the handcrafted version during our quantitative user study, we still argue that our approach outperforms the handcrafted versions in some ways.
While not significant, \netvis~showed overall higher accuracies, which might indicate that the effect between the conditions would become significant, given a larger number of participants.
Furthermore, we saw higher accuracies in key aspects of these types of visualizations, particularly in the direct comparisons such as interpretability and design of the visualization.
Besides that, for the TensorBoard visualizations and original paper figures not all information needed for answering the questions was always present during our quantitative user study.
In such cases, users could answer with -1 whenever information was not available.
Thus, in these cases, users could not give a wrong answer.
We still counted these answers in our evaluation, essentially putting our approach at a disadvantage.
Nonetheless, our techniques outperformed the others in almost all metrics.
Compared to the original paper figures, which ranked better than TensorBoard, our approach has the additional advantage of a unified design and automation, which can save time and makes knowledge transfer possible.
\\
Referring back to the nested model of evaluating visualization design~\cite{munzner2009nested}, first, our evaluation indicates that we indeed work on a relevant problem for our target users.
Moreover, the abstraction level we chose seems to be appropriate as supported by our quantitative user study.
Our expert interviews clarified that it is important to keep such visualizations simple and minimalist as none of the experts complained about missing information in our visualizations.
One expert even explicitly stated that it is important to \emph{emphasize function and architecture especially over obtuse prettiness that you see in many of the tools that visualize activations or things like layer gradients}.
Third, the evaluation of our expert interviews, as well as the quantitative study, suggest that our glyph design is easily interpretable and adds valuable information as it directly visualizes the transformation of data.
Fourth, the interactivity of our implementation shows that we do not have a problem with the speed of the algorithm.
The results of our quantitative user study support this, with an excellent usability score for our system.
While this evaluation is only a first indication of the applicability of~\netvis and only adoption of the concepts in the field can prove its value, the evaluation clearly supports the need for such a tool as well as our design choices.

\section{Conclusion}
In this paper, we propose a visualization design for communicating neural network architectures which is based on expert feedback, state of the art visualizations, and user studies.
Additionally, we provide an automated approach for visualizing CNN architectures and release a data set which contains an analysis of 751 paper figures of neural network architectures from all CVPR and ICCV conference papers since 2013. 
Currently, such visualizations are mostly handcrafted which consumes time in the paper writing process.
Our novel visual grammar for visualizing CNNs, called~\netvis, is informed by an analysis of the current practice, expert feedback, as well as widely accepted data visualization guidelines.
Our proposed visual grammar incorporates visualization requirements for neural network architecture visualization, network layout, aggregation, legend generation, and a novel glyph design.
\netvis~represents the first visualization technique for modern and complex CNNs that is tailored towards use in publications, while the results of our quantitative user study indicate that~\netvis~improves both visualization generation and reading.
For wide adoption,~\netvis~can be used as an online service at \url{https://viscom.net2vis.uni-ulm.de}, where users can obtain CNN architecture visualizations tailored towards use in publications directly from their Keras code.

\ifCLASSOPTIONcompsoc
  \section*{Acknowledgments}
\else
  \section*{Acknowledgment}
\fi

This work was funded by the Carl-Zeiss Scholarship for Ph.D.~students.

\ifCLASSOPTIONcaptionsoff
  \newpage
\fi



\bibliographystyle{IEEEtran}
\bibliography{net2vis}

\vskip -40pt 
\begin{IEEEbiography}[{\includegraphics[width=1in,height=1.25in,clip,keepaspectratio]{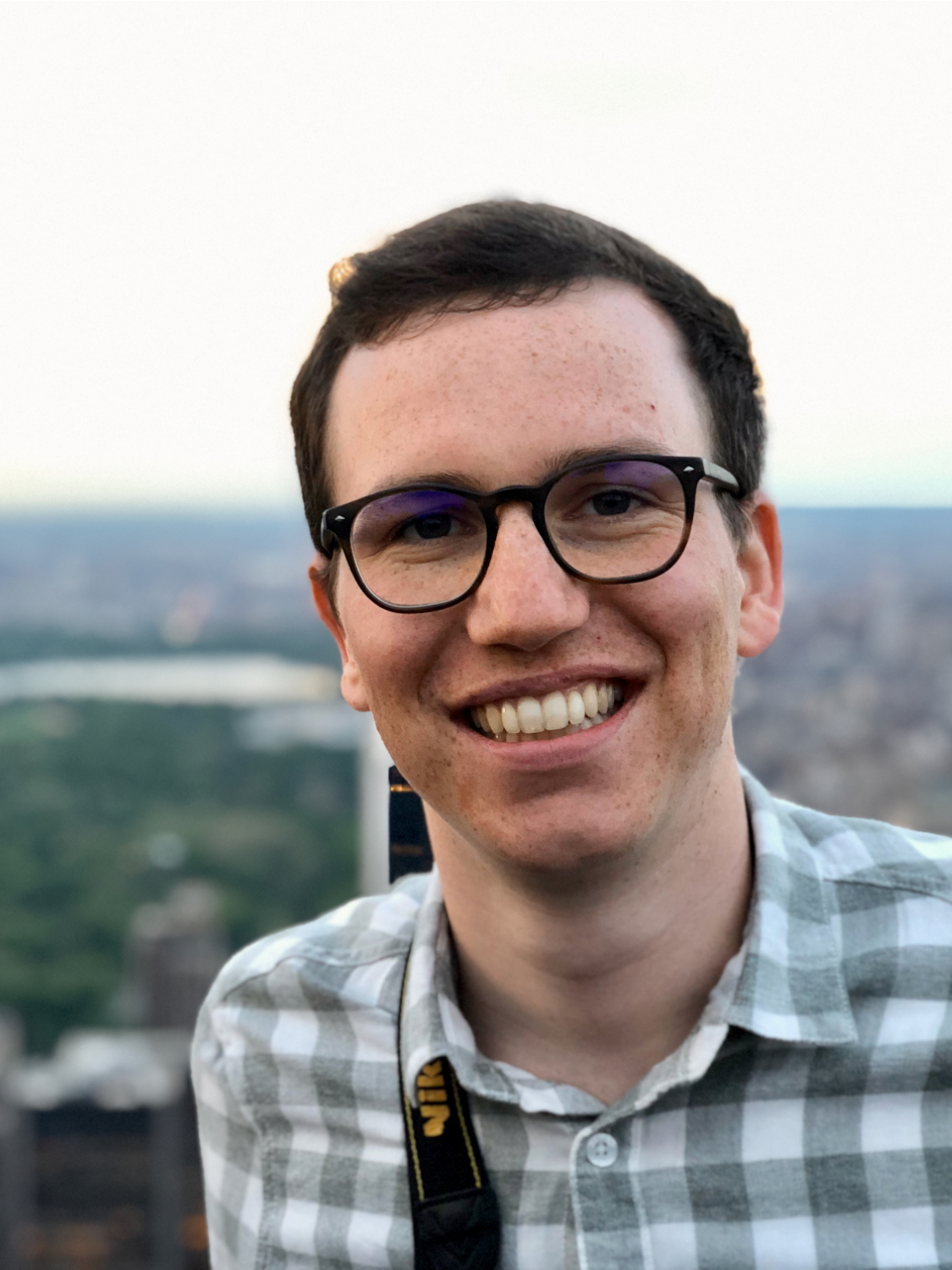}}]{Alex Bäuerle}
  received the master's degree in media informatics from Ulm University in 2017 and is now working at a research associate at the Visual Computing Group at Ulm University.
  His current research interests are the visualization of neural networks to generate better understanding and tooling around these techniques.
\end{IEEEbiography}
\vskip -45pt
\begin{IEEEbiography}[{\includegraphics[width=1in,height=1.25in,clip,keepaspectratio]{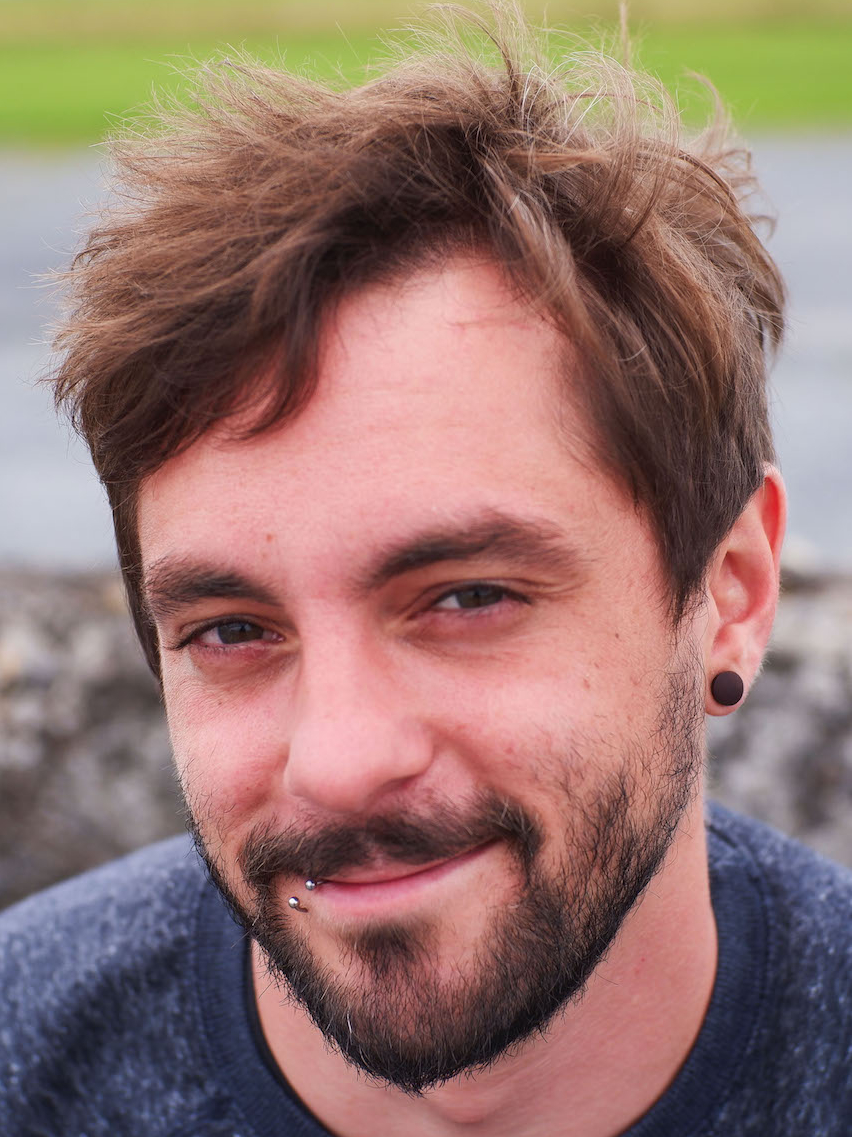}}]{Christian van Onzenoodt}
  received the master's degree in media informatics from Ulm University in 2017 and is now working at a research associate at the Visual Computing Group at Ulm University.
  His current research interests in information visualization with a focus on the perception of visualizations.
\end{IEEEbiography}
\vskip -44pt
\begin{IEEEbiography}[{\includegraphics[width=1in,height=1.25in,clip,keepaspectratio]{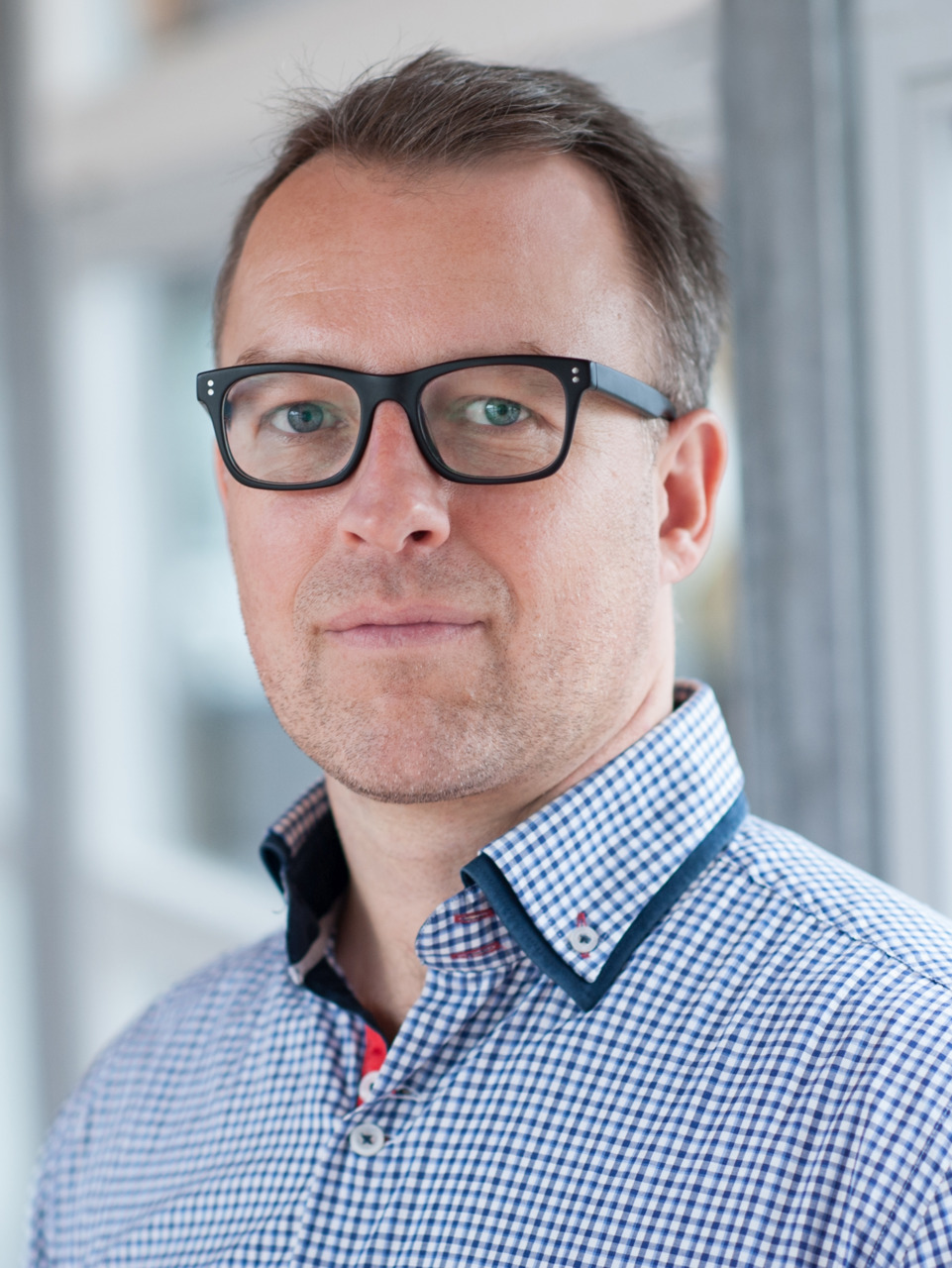}}]{Timo Ropinski}
is a professor at Ulm University, where he is heading the Visual Computing Group.
Before moving to Ulm he was Professor in Interactive Visualization at Linköping University in Sweden, where he was heading the Scientific Visualization Group.
He has received his Ph.D. in computer science in 2004 from the University of Münster, where he has also completed his Habilitation in 2009.
\end{IEEEbiography}

\end{document}